\documentclass[runningheads]{llncs}
\usepackage{xcolor}
\usepackage[linesnumbered,ruled,vlined]{algorithm2e}

\SetCommentSty{mycommfont}
\usepackage{amsmath} 
\usepackage{amsfonts}
\usepackage{hyperref}
\usepackage{float}
\usepackage[bottom]{footmisc}
\usepackage[T1]{fontenc}
\usepackage{graphicx}
\begin{document}

\title{Increasing Rosacea Awareness Among Population Using Deep Learning and Statistical Approaches
\thanks{Accepted to 2024 International Conference on Medical Imaging and Computer-Aided Diagnosis.}
\thanks{This version of the contribution has been accepted for publication, after peer review but is not the Version of Record and does not reflect post-accpetance improvements, or any corrections. The Version of Record is available online soon. Use of this Accepted Version is subject to the publisher's Accepted Manuscript terms of use \url{https://www.springernature.com/gp/open-research/policies/accepted-manuscript-terms}.}}

%
%
\author{Chengyu Yang\text{*} \and Chengjun Liu}
\authorrunning{Chengyu et al.} 
%
%
\institute{New Jersey Institute of Technology, Newark NJ 07103, USA,\\
\email{\{cy322\text{*}, chengjun.liu\}@njit.edu},
\\
WWW home page:\\
\url{https://chengyuyang-njit.github.io/}\\
\url{https://web.njit.edu/~cliu/}
}

\maketitle              
\begin{abstract}

Approximately 16 million Americans suffer from rosacea according to the National Rosacea Society. To increase rosacea awareness, automatic rosacea detection methods using deep learning and explainable statistical approaches are presented in this paper. The deep learning method applies the ResNet-18 for rosacea detection, and the statistical approaches utilize the means of the two classes, namely, the rosacea class vs. the normal class, and the principal component analysis to extract features from the facial images for automatic rosacea detection. The contributions of the proposed methods are three-fold. First, the proposed methods are able to automatically distinguish patients who are suffering from rosacea from people who are clean of this disease. Second, the statistical approaches address the explainability issue that allows doctors and patients to understand and trust the results. And finally, the proposed methods will not only help increase rosacea awareness in the general population but also help remind the patients who suffer from this disease of possible early treatment since rosacea is more treatable at its early stages. The code and data are available at \url{https://github.com/chengyuyang-njit/rosacea_detection.git}.

\end{abstract}
\section{Introduction}



Rosacea is a common skin condition that causes flushing or long-term redness on a person's face, and it may also cause enlarged blood vessels and small, pus-filled bumps \cite{ref_rosacea_def}. Some symptoms may flare for weeks to months and then go away for a while. In its early stages, rosacea might present as mild, transient redness (flushing), which can easily be overlooked. Patients might not seek medical attention until the condition has progressed, which can complicate early diagnosis. Early detection of rosacea, therefore, helps people increase rosacea awareness. Towards that end, we propose automatic rosacea detection methods using deep learning and statistical approaches. 

When applying the supervised deep learning methods on medical images, such as patients' face images suffering from rosacea, there is always a problem of lacking sufficient labelled training data. The confidentiality of patients' information further exacerbates the problem. The fact that some disease is rare makes it even more challenging to collect sufficient training data. 

While research on rosacea and related skin conditions has been conducted using machine learning and computer vision/deep learning algorithms, most high-performing studies leveraging deep learning used datasets with nearly 10,000 images or more. Several studies and others \cite{deep1},\cite{deep2},\cite{deep3},\cite{deep4} utilized substantial data; however, the datasets in these works are fully confidential, making them difficult to reproduce\cite{rosacea_survey}.

As a result, the deep learning methods tend to overfit the limited training data, which often leads to a biased representation of the population distribution. Although there has been work trying to generate more rosacea patients' facial images\cite{ref_rosacea_generate} using the GAN\cite{gan}, no work was done utilizing these generated images to improve the detection of rosasea to the best of our knowledge. Even if in certain cases the deep neutral networks achieve decent performance on the test data, the black-box nature of the deep learning models makes it hard to explain. However, the interpretability and transparency are required to ensure patient safety and clinical acceptance. 

We present a deep learning method and explainable statistical approaches for rosacea detection in this paper to address the issues of both explainable and limitation of the training data. The contributions of our proposed methods are three-fold. First, our proposed methods are able to automatically distinguish patients who are suffering from rosacea from people who are clean of this disease while trained on limited generated data. Second, our statistical approaches address the explainability issue that allows doctors and patients to understand and trust our results. And finally, our proposed methods will not only help increase rosacea awareness in the general population but also help remind the patients who suffer from this disease of possible early treatment since rosacea is more treatable at its early stages.

\section{Automatic Rosacea Detection Using Deep Learning and Statistical Approaches}

We now present our automatic rosacea detection using deep learning and statistical approaches. The deep learning method applies the ResNet-18\cite{ref_resnet} for rosacea detection. The statistical approaches utilize the means of the two classes, namely, the rosacea class vs. the normal class, and the principal component analysis to extract features from the facial images for automatic rosacea detection.

\subsection{Automatic Rosacea Detection Using Deep Learning}
Deep learning has been broadly applied to address various image classification problems. In this paper, we apply the ResNet-18 pretrained on ImageNet\cite{ref_imagenet}. The ResNet-18 method is a convolution neural network (CNN) \cite{ref_cnn} architecture introduced by Kaiming He et al. \cite{ref_resnet}. The ResNet-18 method, which is a lightweight version of the ResNet architecture, is relatively shallow compared to other ResNet variants. The characteristic makes it faster and less computationally expensive yet still achieving a good accuracy. As a result, the ResNet-18 method is broadly used for tasks like image classification.

For automatic rosacea detection, we fine-tune the ResNet-18 model that has been pre-trained on ImageNet. First of all, we fix the weights of all the layers. Then we concatenate a fully-connected layer after the last layer with two outputs, each output corresponds to one class. Finally, we employed the cross entropy loss function to train the model.

\subsection{Automatic Rosacea Detection Using Statistical Approaches}

The first set of statistics includes the means of the two classes in consideration: the rosacea class vs. the normal class. 
In particular, we calculate the mean for both negative and positive samples in the training set.
The mean of a dataset \( \mathbf{x}_1, \mathbf{x}_2, \dots, \mathbf{x}_n \) is given by:
\begin{equation}
    \overline{\mathbf{x}} = \frac{1}{n} \sum_{i=1}^{n} \mathbf{x}_i
\end{equation}
The main idea is to calculate the distance between the test case and negative/positive sample mean respectively and classify the test case to be the one with smaller distance. Here, the distance can be any reasonable distance metric, we use Euclidean distance in our study. This method turns out to be more understandable and relatable to healthcare professionals than the black-box deep neural networks, since this distance metric is a kind of similarity measuring a new patient's similarity with two different groups of population, one is people suffering from rosacea and the others clean of this disease.


The Euclidean distance \(d(\mathbf{x}, \overline{\mathbf{x}})\) between test case \( \mathbf{x} \) and sample mean \( \overline{\mathbf{x}} \) is given by:

\begin{equation}
    d(\mathbf{x}, \overline{\mathbf{x}}) = \sqrt{\sum_{i=1}^{t} (\mathbf{x}_{i} - \overline{\mathbf{x}}_i)^2}
\end{equation}

where \(\mathbf{x}_{i}\) and \(\overline{\mathbf{x}}_{i}\) represent the pixel values of their corresponding images, and \(t\) is the total number of pixels, such as $512 \times 512 \times 3$.


Second, we apply principal component analysis to extract features from the facial images for automatic rosacea detection.
Principal component analysis or PCA is a popular statistical approach for feature extraction \cite{Fukunaga'91}, \cite{Liu'tPAMI00}. Specifically, let $X \in \mathbb{R}^N$ be a random vector representing an image, where $N$ is the dimensionality of the image space.  The covariance matrix of $X$ is defined below:
\begin{equation} \label{eq:CovMatr} 
C_X = E\{[X-E(X)][X-E(X)]^t\}
\end{equation}
where $E(\cdot)$ represents the expectation. The PCA of a random vector $X$ factorizes the covariance matrix ${C}_X$ into the following form:
\begin{equation} \label{eq:PCA} 
C_X = \Phi \Lambda \Phi^t \hspace{0.3cm} with \hspace{0.2cm} \Phi = [\phi_1 \phi_2 \ldots \phi_N], \Lambda = diag\{\lambda_1, \lambda_2, \ldots, \lambda_N\}
\end{equation}
where $\Phi$ is an orthonormal eigenvector matrix and $\Lambda$ is a diagonal eigenvalue matrix with diagonal elements in the decreasing order (${\lambda_1}{\geq}{\lambda_2}{\geq}{\cdots}{\geq}{\lambda_n}$).  $\phi_1, \phi_2, \ldots, \phi_N$ and $\lambda_1, \lambda_2, \ldots, \lambda_N$ are the eigenvectors and the eigenvalues of $C_X$, respectively.

An important property of PCA is the optimal image representation when only a subset of principal components are used to represent the original image \cite{Fukunaga'91}, \cite{Liu'tPAMI00}. Let $P = [\phi_1 \phi_2 \ldots \phi_m]$ where $m < N$ and $P \in \mathbb{R}^{N \times m}$. Applying PCA, we can derive new features as follows:
\begin{equation} \label{eq:DimRed} 
Y = P^t X
\end{equation}
The lower dimensional vector $Y \in \mathbb{R}^m$, therefore, is able to capture the most expressive features of the original data $X$.

Finally, we further improve upon the pixel level Euclidean distance method by applying PCA for feature extraction. The following algorithm shows the detailed procedure for automatic rosacea detection using PCA. 

The detailed algorithm is shown in Algorithm \ref{euclid_pca}.

\begin{algorithm}
    \caption{Automatic Rosacea Detection Using PCA}\label{euclid_pca}

    \DontPrintSemicolon
    \SetAlgoLined
    \SetNoFillComment
    \LinesNotNumbered
    \tcc{$\mathbf{X}_{\mathbf{pos}}$ is the data matrix with each column being a flattened image from rosacea positive class from the training dataset}
    \tcc{$\mathbf{X}_{\mathbf{neg}}$ is the data matrix with each column being a flattened image from rosacea negative class from the training dataset}
    \tcc{$\mathbf{x}$ is the flattened test image}
    \tcc{$\mathbf{r}$ is  the number of principal components selected in the PCA}
    \KwData{{$\mathbf{X_{pos}}, \mathbf{X_{neg}},\mathbf{x}, \mathbf{r} $}}
    \;
    \tcc{returned 0 means the the test image is rosacea negative}
    \tcc{returned 1 means the test image is rosacea positive}
    \KwResult{0 or 1}
    \;
    \tcc{Get the mean image for both the rosacea positive and negative samples}
    $\overline{\mathbf{x}}_{\mathbf{pos}} = \mathbf{mean}(\mathbf{X}_{\mathbf{pos}}) $\;
    $\overline{\mathbf{x}}_{\mathbf{neg}} = \mathbf{mean}(\mathbf{X}_{\mathbf{neg}}) $\;
    \;
    \tcc{Do the PCA on the training data using singular value decomposition(SVD) respetively}
    $\mathbf{U}_{\mathbf{pos}},\mathbf{S}_{\mathbf{pos}},\mathbf{V}_{\mathbf{pos}}^{\mathbf{T}} = \mathbf{SVD}(\mathbf{X}_{\mathbf{pos}})$ \;
    $\mathbf{U}_{\mathbf{neg}},\mathbf{S}_{\mathbf{neg}},\mathbf{V}_{\mathbf{neg}}^{\mathbf{T}} = \mathbf{SVD}(\mathbf{X}_{\mathbf{neg}})$ 
    
    \;
    
    \tcc{Get first r principal components}
    $\mathbf{U_{pos}} \gets \mathbf{U_{pos}[:r]}$\;
    $\mathbf{U_{neg}} \gets \mathbf{U_{neg}[:r]}$
    
    \;
    \tcc{Project the mean images obtained above onto the row space of principal components matrix}
    $\overline{\mathbf{x}}_{\mathbf{pos}}' = \overline{\mathbf{x}}_{\mathbf{pos}} \cdot \mathbf{U_{pos}[:r]}$\;
    $\overline{\mathbf{x}}_{\mathbf{neg}}' = \overline{\mathbf{x}}_{\mathbf{neg}} \cdot \mathbf{U_{neg}[:r]}$
    
    \;
    \tcc{Project the test image to the row space of principal component matrices of the two classes respectively}
    $\mathbf{x'_{pos}} = \mathbf{x} \cdot \mathbf{U_{pos}[:r]}$\;
    $\mathbf{x'_{neg}} = \mathbf{x} \cdot \mathbf{U_{neg}[:r]}$

    \;
    \tcc{Compare the projected images' distances and get the result}
    \eIf{$||\mathbf{x'_{pos}} - \overline{\mathbf{x}}_{\mathbf{pos}}'||^2 < ||\mathbf{x'_{neg}} - \overline{\mathbf{x}}_{\mathbf{neg}}'||^2$}
        {return 1}
     {return 0}

\end{algorithm}

\section{Experiments}

We first build a training and validation data set using the generated images \cite{ref_rosacea_generate}, \cite{ref_style_gan} to train the proposed deep learning and statistical approaches. We then create a test data set using real images to evaluate the performance of our proposed methods.

\subsection{Datasets}

Two data sets are presented in this paper: the training and validation data set and the test data set, which are detailed below. 

\subsubsection{Training and Validation Data Set}
\begin{figure}
\includegraphics[width=\textwidth]{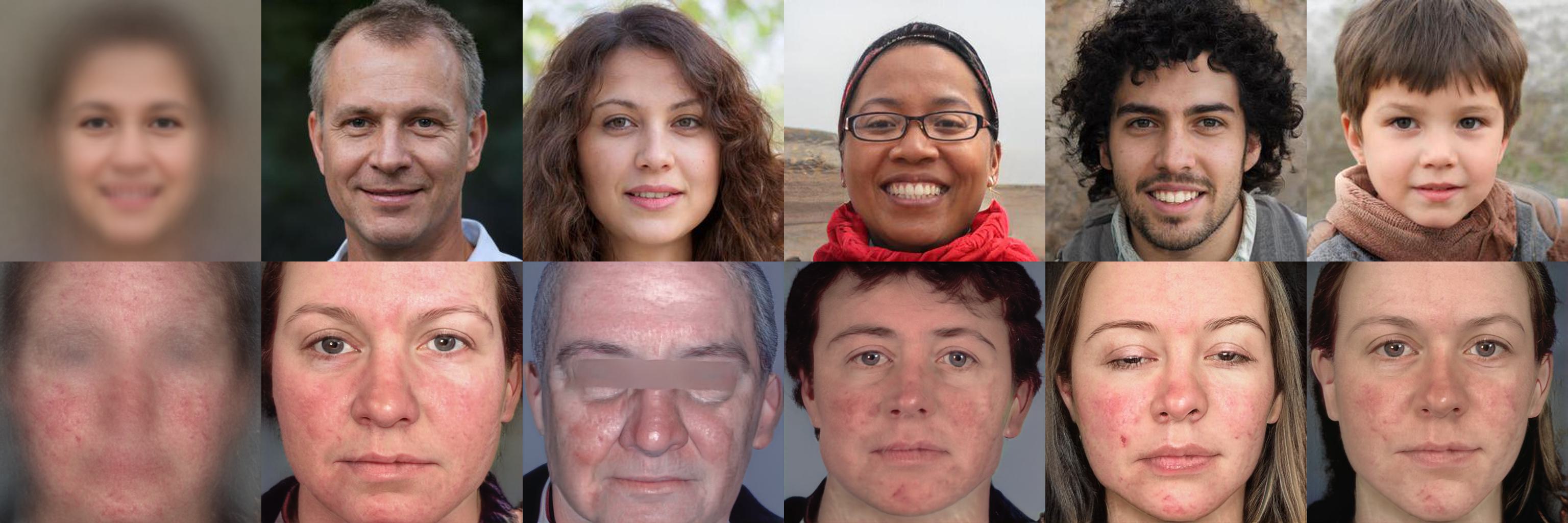}
\caption{The first column shows the mean images of the two datasets from the rosacea negative and positive classes, respectively. The remaining five columns display five random example images from the two datasets corresponding to the normal people and those with rosacea, respectively.} \label{fig1}
\end{figure}

In order to simulate the real situation, that is, the amount of training data is limited and hard to collect, and also with the purpose of protecting patients' privacy, we use generated images for both rosacea positive and negative cases. As is shown in Fig\ref{fig1}, the frontal face images of patients with rosacea have been generated using the GAN \cite{ref_rosacea_generate}\cite{gan}. We therefore use 300 of the generated rosacea frontal face images and split them into the training set with 250 images and the validation set with the remaining 50 images. For the rosacea negative cases, we use 600 frontal face images generated from the Style-GAN \cite{ref_style_gan} and split them into a training set with 500 images and the validation set with the remaining 100 images. 

\subsubsection{Test Data Set}
To assess the performance of the proposed deep learning and statistical approaches on real images, we collect 50 real frontal face images with rosacea from various sources including \href{https://www.kaggle.com/}{Kaggle}, \href{https://dermnetnz.org/}{DermNet}, and \href{https://www.rosacea.org/}{National Rosacea Socity}. These images are selected with a standard of having at least 200 x 200 pixel resolution. The face in each image is detected and aligned using the DeepFace\cite{ref_df} and cropped and resized to the same size with the training and the validation images which are $512 \times 512 \times 3 $ for the purpose of testing. In addition, we include 150 real people's frontal faces not having rasacea from the CelebA Dataset\cite{celeba} in our test data set.

\subsection{Automatic Rosacea Detection Using Deep Learning Method}
We employ the ResNet-18 pretrained on ImageNet. All the previous layers are fixed except the last one, which is a fully-connected layer with two outputs corresponding to two classes, one is rosacea positive, the other negative. The loss function is cross entropy.  The training is finished after 50 epochs, with batch size of 4. And the Stochastic Gradient Descent (SGD) is employed for the training with a learning rate of 0.001 and the momentum being 0.9.
\begin{figure}[!ht]
\includegraphics[width=\textwidth]{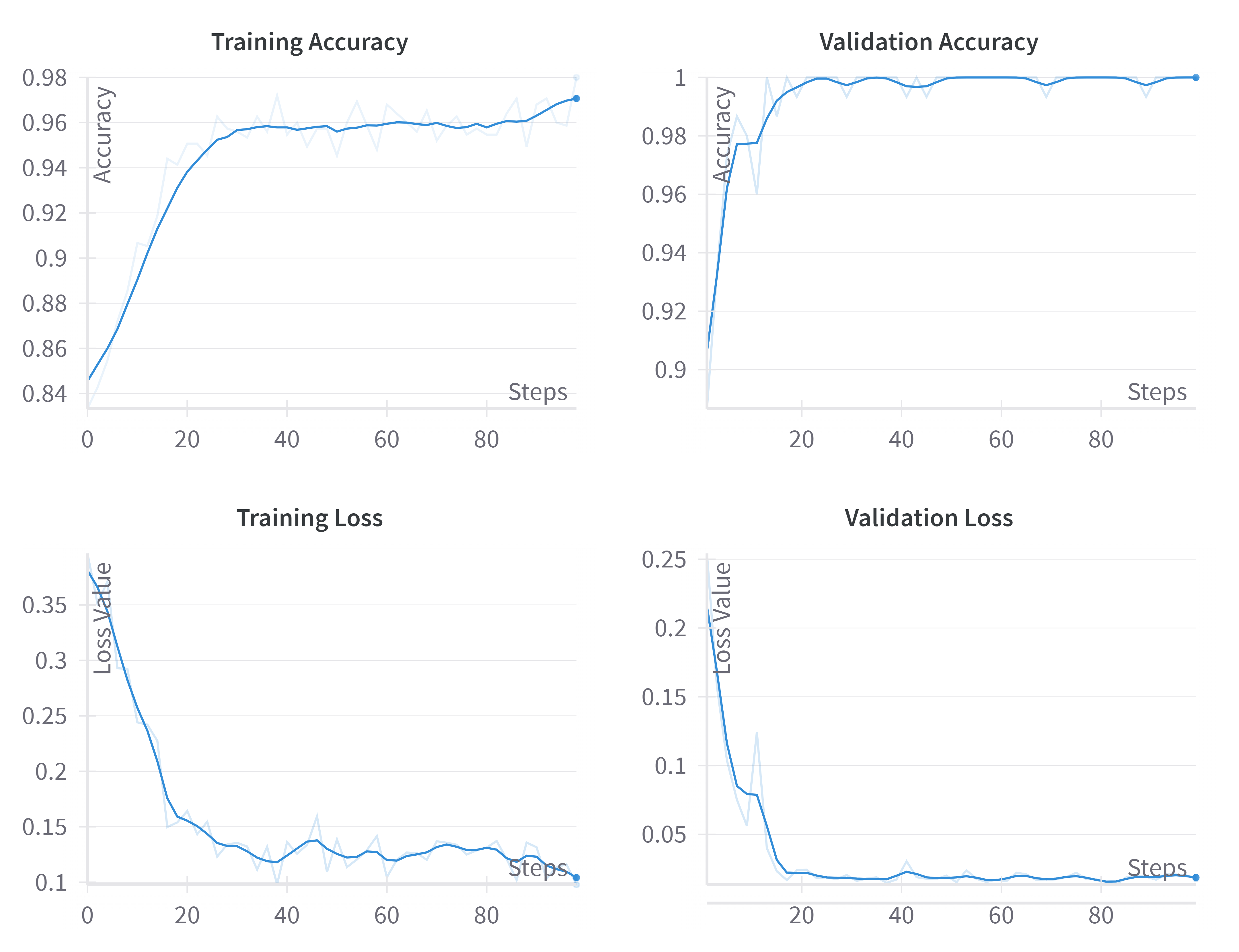}
\caption{The training/validation loss and accuracy of the ResNet-18.} \label{fig:image3}
\end{figure}
As is shown in \textbf{fig.}\ref{fig:image3}, the validation accuracy has achieved 100\%. 

\begin{table}\caption{Confusion matrix for the deep learning method on the test set}\label{table1}
\centering
$\begin{array}{c|c|c}
\hline
  & \textbf{Predicted Rosacea} & \textbf{Predicted Not Rosacea} \\
\hline
\textbf{Rosasea} & 29 & 21 \\
\hline
\textbf{Not Rosasea} & 0 & 150 \\
\hline
\end{array}$
\end{table}

Table \ref{table1} shows that even though the deep learning achieves 100\% accuracy on the validation set, due to the limitation on the size and the generated nature of the training samples, the deep learning method suffers from the problem of overfitting and cannot generalize well to the real world test dataset.  Actually, \textbf{Tables} \ref{table1}  and \ref{table4} show that the recall rate is only 0.58 and 21 positive cases are reported to be negative. In this case, a lot of patients actually with rosacea will not be diagnosed and will miss the best opportunity to start early treatment. 
Another drawback of the deep learning method is due to its black-box nature, which makes it hard to explain the reason for the diagnosis.

\subsection{Automatic Rosacea Detection Using Statistical Approaches}

To assess the performance of our proposed statistical approaches on the real image from the test data set, we set r to 180 as the hyperparameter in Algorithm \ref{euclid_pca}. Specifically, Table \ref{table2} shows the automatic rosacea detection performance of the proposed deep learning and the statistical approaches, respectively, on the validation data set.

\begin{table}[H]\caption{The automatic rosacea detection performance on the validation data set of the proposed deep learning and the statistical approaches, respectively.}\label{table2}
\centering
$\begin{array}{c|c|c|c|c}
\hline
  & \textbf{Accuracy} & \textbf{Precision} & \textbf{Recall} & \textbf{F1 Score} \\
\hline
\textbf{ResNet-18} & 1.00 & 1.00 &1.00 &1.00\\
\hline
\textbf{Stats Method with PCA}&0.90  &0.77  & 0.96 & 0.86\\
\hline
\end{array}$
\end{table}

\vspace{-2em}

\begin{table}[H]\caption{The confusion matrix for the statistical method on the test data set with PCA}\label{table3}
\centering
$\begin{array}{c|c|c}
\hline
  & \textbf{Predicted Rosacea} & \textbf{Predicted Not Rosacea} \\
\hline
\textbf{Rosacea} & 44 & 6 \\
\hline
\textbf{Not Rosacea} & 15 & 135 \\
\hline
\end{array}$
\end{table}

\vspace{-2em}

\begin{table}[H]\caption{The automatic rosacea detection performance on the test data set using the proposed deep learning and the statistical approaches, respectively.}\label{table4}
\centering
$\begin{array}{c|c|c|c|c}
\hline
  & \textbf{Accuracy} & \textbf{Precision} & \textbf{Recall} & \textbf{F1 Score} \\
\hline
\textbf{ResNet-18} & \textbf{0.895} & \textbf{1.00} & 0.58 &0.73\\
\hline
\textbf{Stats Method with PCA}&\textbf{0.895} & 0.75& \textbf{0.88}& \textbf{0.81}\\
\hline
\end{array}$
\end{table}

The confusion matrix in \textbf{Table} \ref{table3} reveals that only 6 patients who actually suffered from rosacea are not detected by the statistical approach, which is a great improvement upon the ResNet-18 deep learning method with 21 patients suffering with rosacea that are not diagnosed. \textbf{Table} \ref{table4} shows that with the same accuracy and relatively lower precision rate, the statistical approach with PCA achieves a higher overall F1 score and a much higher recall rate than the deep learning method. A higher recall rate will not only help increase rosacea awareness in the general population but also help remind the patients who suffer from this disease of possible early treatment since rosacea is more treatable at its early stages.

\section{Conclusion}
In this paper, we present a study on automatic rosacea detection using both deep learning and statistical approaches with an objective of increasing the awareness of this skin condition among the general population. We use the ResNet-18 architecture for deep learning and apply statistical approaches including the principal component analysis (PCA) to extract features from facial images for automatic rosacea detection. Our findings indicate that while the deep learning method achieves high accuracy, it struggles with overfitting and does not generalize well to the real world data. Therefore, the deep learning method can not detect the existence of rosacea in a number of real image cases. It also lacks explainability due to its black-box nature, which is crucial for clinical acceptance. In contrast, the statistical methods demonstrate much better recall rates, effectively identifying patients with rosacea and addressing the interpretability issue by measuring the distance metric between the test case and the sample mean. The statistical methods are designed to be understandable and relatable to healthcare professionals. By focusing on explainable results, the statistical methods facilitate discussions between doctors and patients regarding diagnosis and treatment options, thereby fostering trust and confidence in the results. 

\bibliographystyle{splncs04}


\end{document}